\documentclass[conference]{IEEEtran}
\usepackage{amsmath}
\usepackage[usenames, dvipsnames]{color}
\usepackage{cite}
\usepackage{graphicx}
\usepackage{verbatim}
\usepackage{subcaption}
\usepackage{url}

\graphicspath{{/Figures/pdf/}{/Figures/jpg/}{/Figures/png/}}

\begin{document}
\title{An error-propagation spiking neural network compatible with neuromorphic processors} 

\author{\IEEEauthorblockN{Matteo Cartiglia,
Germain Haessig and
Giacomo Indiveri}
\IEEEauthorblockA{Institute of Neuroinformatics\\
University of Zurich and ETH Zurich,
Zurich, Switzerland}
Email: \url{[camatteo, germain, giacomo]@ini.uzh.ch}}

\maketitle

\begin{abstract}

Spiking neural networks have shown great promise for the design of low-power sensory-processing and edge-computing hardware platforms.
However, implementing on-chip learning algorithms on such architectures is still an open challenge, especially for multi-layer networks that rely on the back-propagation  algorithm. In this paper, we present a spike-based learning method that approximates back-propagation using local weight update mechanisms and which is compatible with mixed-signal analog/digital neuromorphic circuits. We introduce a network architecture that enables synaptic weight update mechanisms to back-propagate error signals across layers and present a network that can be trained to distinguish between two spike-based patterns that have identical mean firing rates, but different spike-timings.
This work represents a first step towards the design of ultra-low power mixed-signal neuromorphic processing systems with on-chip learning circuits that can be trained to recognize different spatio-temporal patterns of spiking activity (e.g. produced by event-based vision or auditory sensors).

%
%
%
%
%

\end{abstract}

\section{Introduction}

Spiking neural network (SNN) models are highly efficient at processing sensory signals while minimizing their memory and power-consumption resources. For this reason, they have established themselves as a valuable alternative to traditional deep learning methods for applications that require ultra-low power edge-computing capabilities. These models, compared to standard Artificial Neural Networks (ANNs), take one step closer to brain-inspired processing by using an event-driven processing mode: their leaky Integrate-and-Fire (I\&F) neurons transmit information only when there is sufficient input data to reach their spiking threshold. As the I\&F neurons are implemented with passive current-mode sub-threshold circuits~\cite{Chicca_etal14}, this data-driven computation mode only burns power when there are signals to process. However, training neural networks implemented with these circuits remains a challenging problem due to their intrinsic spiking non-linearity and to the requirement of local learning rules. To this end, adapting the original back-propagation algorithm~\cite{Rumelhart_etal86a} to work with spiking neural networks has drawn substantial attention~\cite{Lee_etal16,Neftci_etal17,Rueckauer_etal17}. So far, however, these attempts have been done for software simulations of spiking neurons and the underlying incongruity between the proposed training mechanism and the limitations of the (noisy, imprecise and low-resolution) mixed-signal analog/digital silicon neurons remains. 

Dedicated neuromorphic spiking neural network platforms able to implement learning mechanism have indeed been proposed in recent years~\cite{Schemmel_etal08, Frenkel_etal18, Merolla_etal14a, Davies_etal18}. However, such architectures have been designed with pure digital circuits, or with digital co-processors in the loop. They have not been optimized for power consumption, hence rendering the use of such networks less efficient for low power edge computing applications. 

The contribution of this paper, is to investigate the feasibility of creating a dedicated mixed-signal learning circuit that implements a spiking version of a recently proposed cortical model~\cite{Sacramento_etal18} which approximates the back-propagation algorithm through the propagation of local error signals at the network level. We show that this network is able to correctly learn to distinguish different temporal input patterns, relying solely on feedback propagation of a teacher stimulus and local weight updates mechanism, compatible with mixed-signal neuromorphic circuits~\cite{Chicca_etal14}.
The main challenge in this work was to find a spiking neural network architecture that is able to approximate the back-propagation algorithm and at the same time is compatible with mixed-signal learning circuits.

\section{Network topology}

\begin{figure}
\centering
\includegraphics[width=0.4\textwidth]{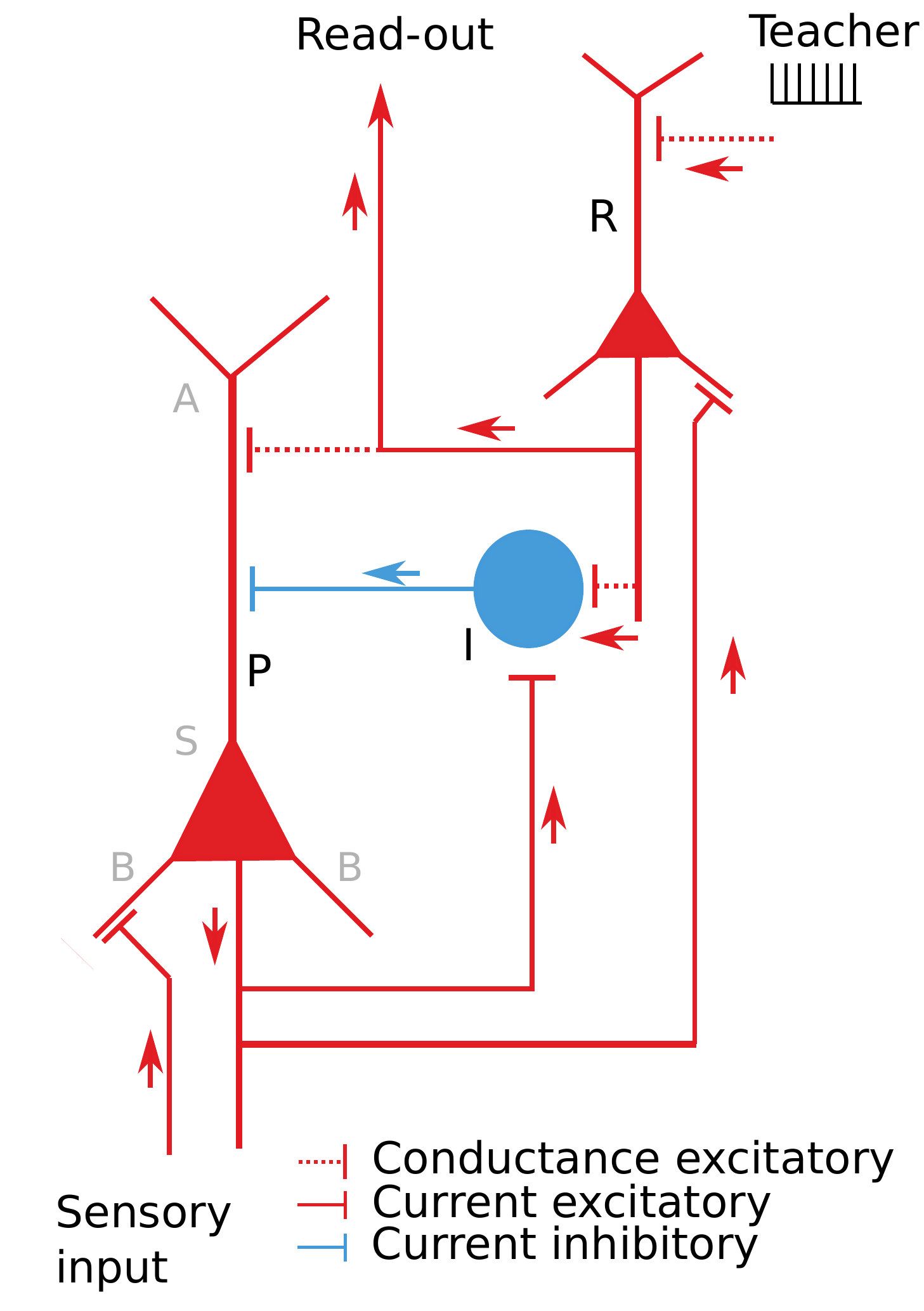}
\caption{Network connectivity between (P)yramidal, (I)nter and (R)eadout neurons. Sensory input is propagated bottom-up via the pyramidal neuron; the teacher signal is propagated top-down via the readout neuron. Recurrent connections between pyramidal neurons and interneurons are set-up to minimize the prediction error in the apical compartment of the pyramidal neuron. Dotted connection are conductance-base excitatory synapses, as described in Section~\ref{sec::equations}}
\label{fig:Simple_network}
\end{figure}

Inspired by the functionality, connectivity, and diversity of cell types found in the neocortex, previous work~\cite{Sacramento_etal18} has proposed a cortical model of learning and neural computation comprising of a population of rate-based multi-compartment excitatory neurons and inhibitory interneurons. Multi-compartment neurons aim to approximate dendritic processes more accurately by having separate compartments, which interact with each other.
Figure~\ref{fig:Simple_network} shows a simplified representation of such a model: the neuron is schematized as a three-compartment ((A) Apical, (B) Basal and (S) Somatic compartment) pyramidal neuron (P) which integrates sensory bottom-up information and a top-down teaching signal. Pyramidal neurons are driven by sensory inputs, while interneurons (I) are laterally driven by the pyramidal neurons. The apical compartment (A) of the pyramidal neurons receive feedback from higher-order areas as well as from the interneurons. 

All bottom-up connections, as well as lateral recurrent connections, (see Figure~\ref{fig:Simple_network}) are subject to plasticity, while the synapses that modulate the top-down teaching signals are fixed and conductance-based (see Figure~\ref{fig:Simple_network}, dotted connections, Eq.~\ref{eq4} and~\cite{Bartolozzi_Indiveri07a}). Plasticity is driven by the requirement of the apical compartment to match the top-down teaching signal coming from the layer above with the lateral connectivity coming from the interneurons. This requires the inhibitory interneuron to match the activity of the above layer. As such the hidden layer has a number of interneurons equal to the number of neurons in the layer above.

During learning, interneurons learn to replicate the spiking activity of specific neurons in the above layer, such as to cancel the top-down teacher signal in the apical compartment of the pyramidal neuron~\cite{Urbanczik_Senn14}.
Similarly, the firing rate of the readout neuron (R) will tend to the teacher signal.
The weights of the network converge so that the firing rate of the readout neuron will tend to the teacher signal, even when the teacher signal is turned off.

\section{Constraints and models for the hardware implementation}
\subsection{Equations}
\label{sec::equations}

In the following, we adapted the model to use equations and plasticity mechanisms that can be directly implemented with sub-threshold neuromorphic circuits as presented in~\cite{Chicca_etal14}, and that are implemented on the DYNAP-SE chip~\cite{Moradi_etal18}. 

In the theoretical model, the voltage $V_{Comp}^k$ of individual compartments is approximated by low-pass filtering the incoming current $I_k$ of the $k$--th compartment (Eq.~\ref{eq1}). For compatibility with the current-mode neuromorphic circuits, the membrane voltage, is represented as a current $I_S$, as defined in Eq.~\ref{eq2}. This membrane \textit{current} will trigger an action potential when crossing a predefined threshold.

\begin{equation} 
I_{Comp}^k = - \sum \limits_i \frac{d I_i}{d t}
\label{eq1}
\end{equation}

\begin{equation}
\cfrac{ d I_{S}}{d t} = \frac{1}{\tau}\sum \limits_{k} \frac{\alpha_{k}}{\alpha_{k} + l}  I_{Comp}^k- l  I_{S} + \sigma^2
\label{eq2}
\end{equation}
where $\tau$ is the membrane time constant, $\alpha_k$ the coupling factor of the $k$--th compartment, $l$ the leak and $\sigma^2$ an additional noise term (see Section~\ref{sec::mismatch}).

Bottom-up weight updates (Figure~\ref{fig:Simple_network}, arrowed connections) are modulated by the learning rule in Eq.~\ref{eq3}. The postscripts S and B represent different compartments of the neuron (such as the somatic and basal compartments), but the same learning rule takes place in the (A) apical compartment of the pyramidal neuron between the top-down projection and the lateral inhibitory connection. The hysteresis term ($\Theta$) of the equation gives rise to a stop-learning region which enables life long learning~\cite{Payvand_Indiveri19}.

Bottom-up connections follow the following learning rule :
\begin{equation}
\Delta w =  
\begin{cases} 
\eta   (I_{S}-I_{B})-\Theta  \hspace{0.3cm} & \text{if } I_{S}-I_{B} > \Theta\\
\eta   (I_{S}-I_{B})+\Theta  \hspace{0.3cm} & \text{if } I_{S}-I_{B} < -\Theta\\
0 & \text{else}
\end{cases}
\label{eq3}
\end{equation}
where $\Delta w$ is the incremental weight update to be applied, being a function of the difference between somatic and basal membrane currents $I_S$ and $I_B$, $\eta$ a learning update rate and $\Theta$ an hysteresis term. 

Top-down conductance-based connections (Figure~\ref{fig:Simple_network}, dotted connections) are described by Eq.~\ref{eq4}. 
\begin{equation}
I_{Comp}^{A} = I_{syn}\alpha(I_{S}- E_{rev})
\label{eq4}
\end{equation}

Where $I_{Comp}^{A}$ is the apical compartment current injected into the somatic compartment, $I_{S}$ is the somatic membrane current and $E_{rev}$ is the reversal potential.  Finally, $\alpha$ is a scaling factor and $I_{syn}$ is the post-synaptic current.				

\subsection{Parameters mismatch}
\label{sec::mismatch}
This work was carried out using the Python Teili software library~\cite{Milde_etal18} for the Brian spiking neural network simulator~\cite{Goodman_Brette08} which reduces the gap between software simulation and hardware emulation by directly simulating the properties of the mixed-signals CMOS neuron circuits. In order to model the device-mismatch effects due to the fabrication process of circuits, all the variables in the neuron and synapse network are subject to a 20\% random variability, centered around their nominal value, which is in the range of the expected variations \cite{Serrano-Gotarredona_Linares-Barranco99, Pavasovic_etal94}.

\subsection{Learning Circuits}
\label{sec::learningCircuits}

The learning rules described in section~\ref{sec::equations} have already been implemented in previously proposed neuromorphic hardware~\cite{Qiao_etal15}.
Conductance based synapses, such as the top-down connections in this work, have been described and tested thoroughly~\cite{Bartolozzi_Indiveri07a}. Using the NMDA block of single neurons, these equations are implementable on mixed-signal neuromorphic processors and have been used in spiking dendritic prediction algorithms~\cite{Sumislawska_etal16}.
The hardware implementation of the bottom-up connections have been implemented in recent learning circuits~\cite{Payvand_etal18} and can be extended to exploit the stochasticity of memristive devices.

\begin{figure}
\centering
\includegraphics[width=0.48\textwidth]{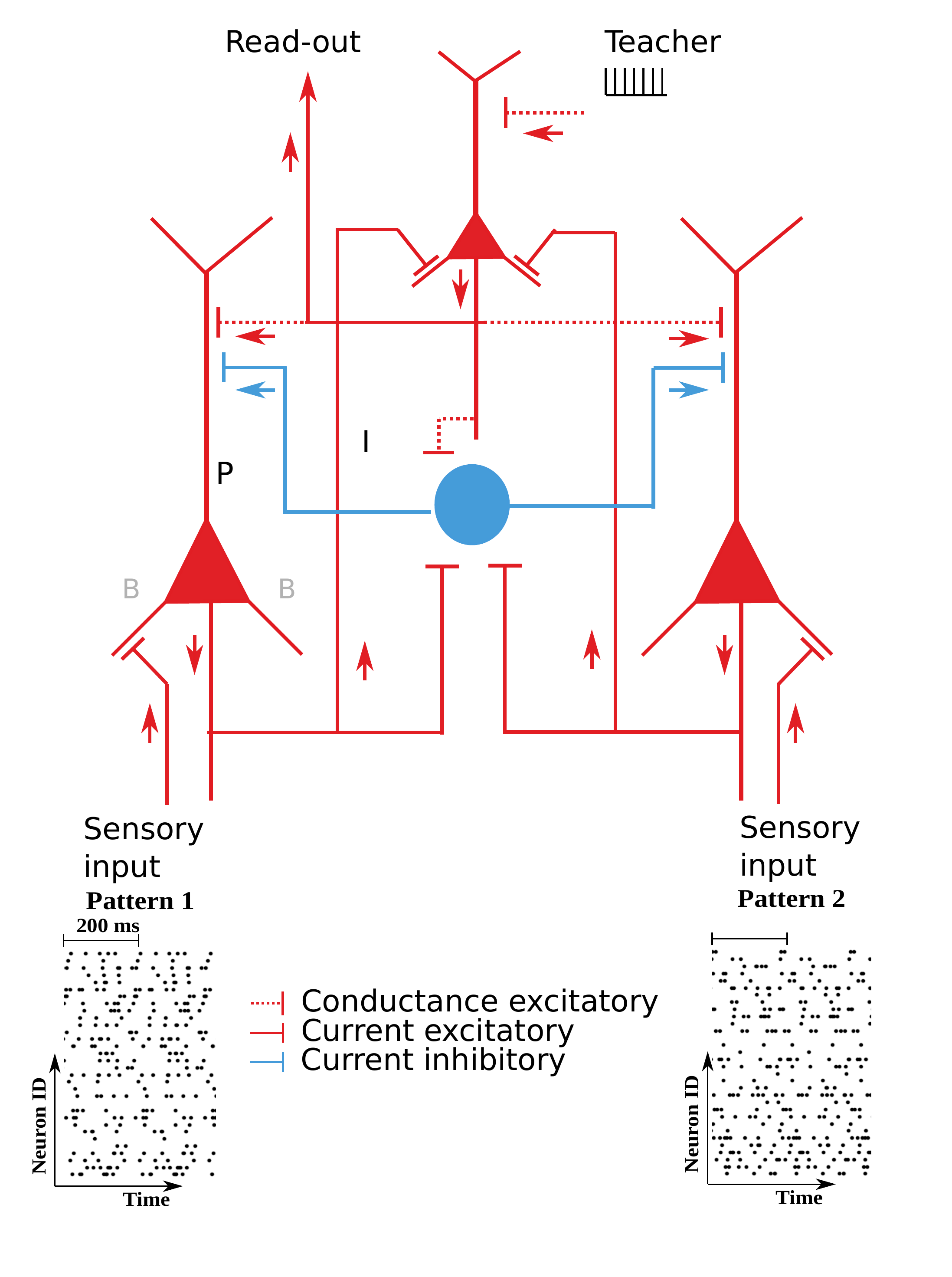}
\caption{Architecture of the network. The two input patterns have identical statistics: 128 spikes for a 200ms pattern, distributed on 32 input neurons.}
\label{fig:Pattern_recognition}
\end{figure}

\begin{figure}
\centering
\includegraphics[width=0.45\textwidth]{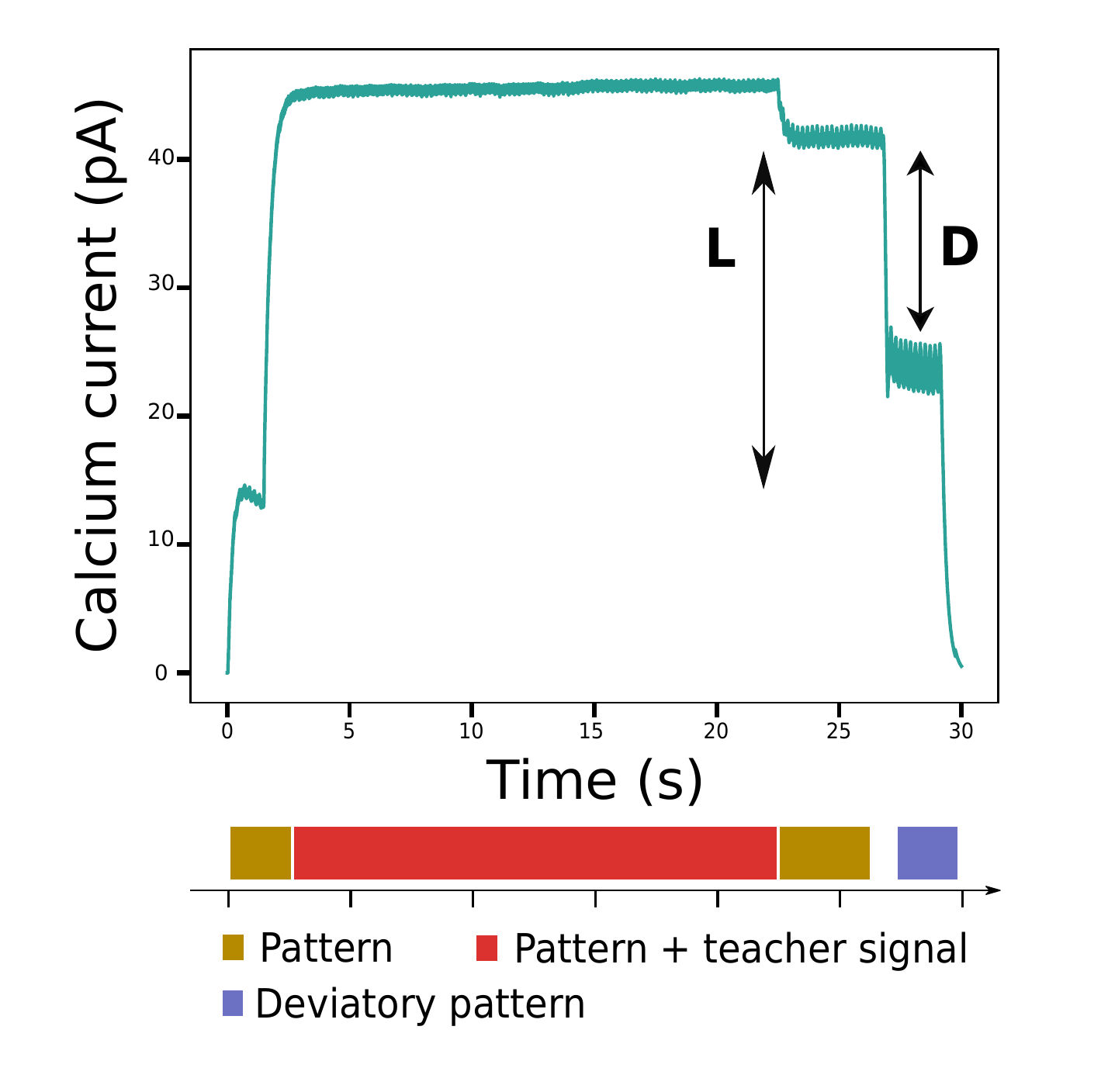}
\caption{Results from the pattern recognition task. The plot shows the calcium current which is approximated by the low-passed (200ms time constant) spiking activity of the readout neuron. Spikes are not shown due to their oscillatory dynamics.
The amplitude L represents the learning process: the difference in activity of the output neuron between before and after the  presentation of teaching signal and input pattern. The amplitude D represents the difference in activity between a known pattern (Pattern 1) and a deviatory sensory input (Pattern 2), proving that learning is specific. Optimally, we would expect L and D to be as large as possible.
The spiking activity of the output neuron is measured as a current due to hardware constraints as explained in~\ref{sec::equations}.
}
\label{fig:Pattern_recognition_result}
\end{figure}

\section{Results}
Following the rate-based model previously proposed~\cite{Sacramento_etal18}, we validate the spiking network in a spatio-temporal pattern recognition task: two distinct spatio-temporal patterns, with same mean firing rates, are presented to the network, which was trained to respond only to one of the two. A network comprising two pyramidal neurons and one interneuron is able to learn to detect a 32 channel input sensory pattern. Figure~\ref{fig:Pattern_recognition} describes the proposed network.
A synthetic randomly generated periodic spiking pattern was presented together with a 900~Hz teaching signal. The teacher signal was presented for an arbitrary long period during which the network learns to associate the higher firing activity due to the teaching signal with an input pattern and nudges the weights in order to sustain this activity even when the teaching signal is turned off. It is then able to differentiate the input pattern (Pattern 1) from a deviating pattern (Pattern 2) with identical statistics.

Figure~\ref{fig:Pattern_recognition_result} shows the results of the pattern recognition task: first, the input pattern is presented without the teaching signal, then the teaching signal and the input pattern are presented together. Finally, the teaching signal is removed. The activity of the output neuron has increased by a factor L  due to learning process that took place. After a period in which no input is presented, a second, deviatory pattern is presented to the network: the activity of the output neuron is lower by a factor of D. The response of the untrained network to either pattern is comparable, signifying that the learning process is selective to the shown pattern.

\begin{figure}
 \centering
\includegraphics[width=0.4\textwidth]{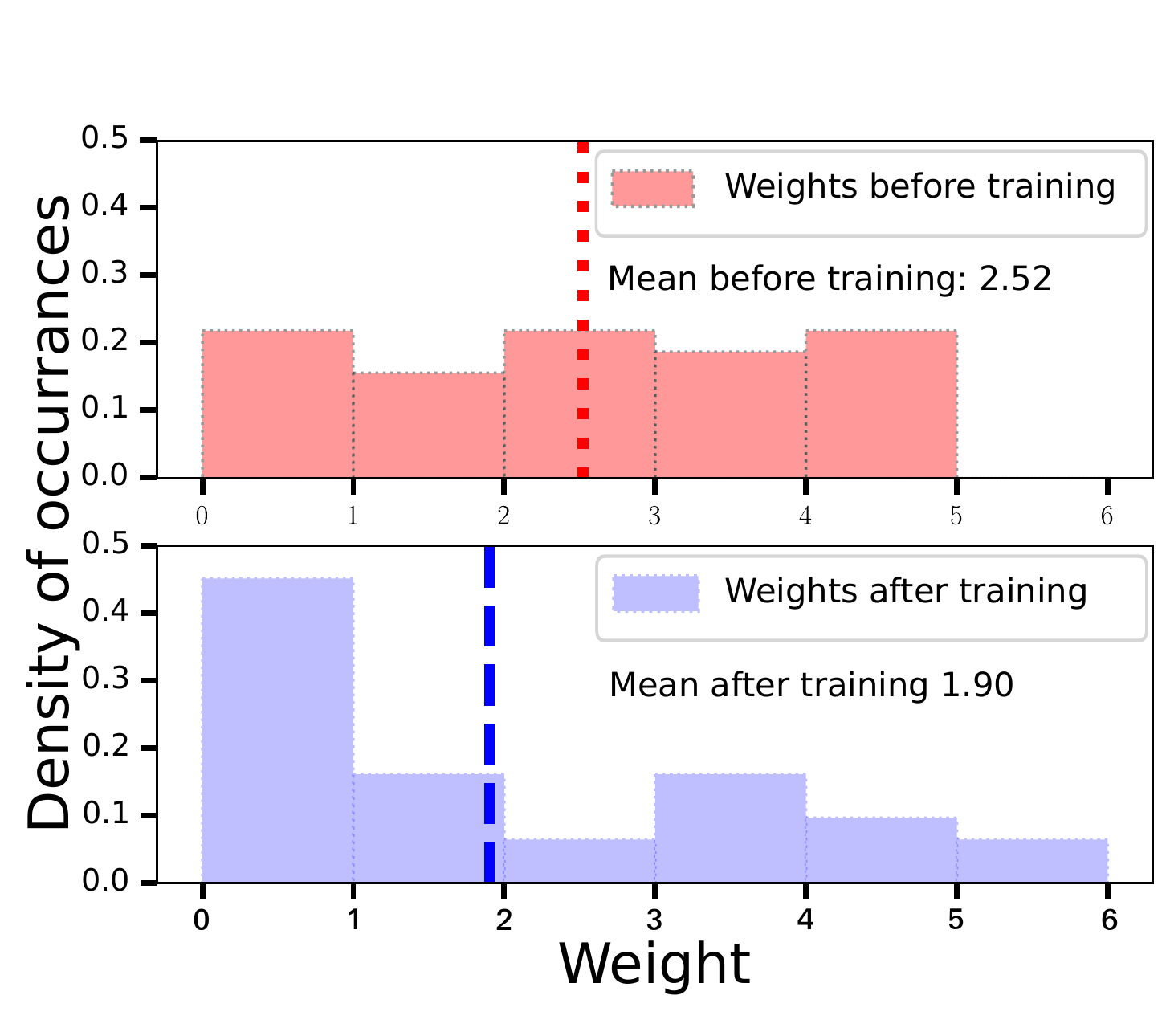}
\caption{Input weight distribution before (in red, top) and after (in blue, bottom) learning. The density of occurrences indicates the probability density distribution of the weights. Weights are initialized randomly from a uniform distribution and can evolve freely. The different distributions proves that after learning the network becomes selective to specific inputs. }
\label{fig:input_Weight_evolution}
\end{figure}


Figure~\ref{fig:input_Weight_evolution} shows how the input weights get modified during the training phase. The different distribution of the input weights between before and after learning proves that the network becomes selective for specific inputs that, combined, enable the output neuron to match the desired target. 

\begin{figure}
\centering
\includegraphics[width=0.45\textwidth]{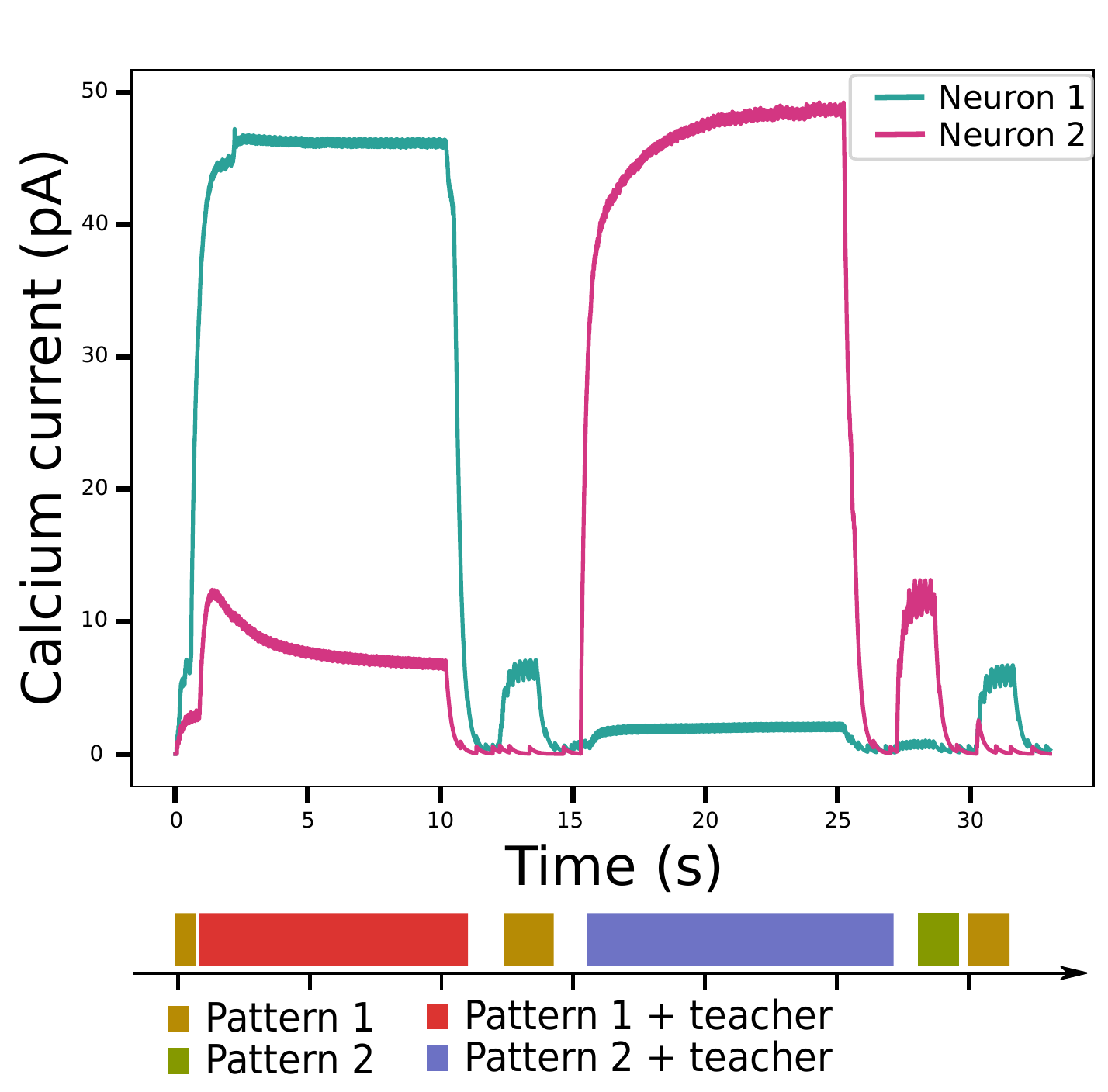}

\caption{Results of the pattern discrimination task. The Calcium current plotted in the figure represents the average firing rate (low-pass filtered with a time constant of 200\,ms) of the readout neuron. After learning, when the teacher signal is turned off, the network can correctly distinguish the two patterns. During classification, the neuron that encodes the pattern which is not presented is silent. }
\label{fig:Pattern_separation_result}
\end{figure}

\subsection{Pattern discrimination}
Building on the results of the pattern recognition task, described above, we improved the network to be able to correctly classify separate temporal patterns. The hidden layer of the network has two interneurons and 8 pyramidal neurons. The teaching signal, as well as the two input patterns, are kept the same between experiments.

Figure~\ref{fig:Pattern_separation_result} shows the results of the pattern discrimination task. Each sensory pattern is fed in the presence of the teaching signal, specifically only the output neuron that is attributed in classifying the input pattern is stimulated. After having trained the network on both patterns, each pattern is presented in the absence of the teaching signal. The spiking activity of the output neurons, correctly classify the input patterns.
In order to optimize the hardware resources at our disposal (limited number of synapses on neuromorphic chips~\cite{Moradi_etal18}), synaptic connections were created with 50\% sparsity constraint: each sensory neuron has a 50\% probability of being connected to any neuron in the hidden layer and analogously the neurons in the hidden layer have 50\% probability of being connected to any neuron in the readout layer. Weights are randomly initialized with uniform distributions, and can freely evolve. It should be noted that the teacher signal is presented for the same amount of time for both patterns.

\section{Conclusion}

In this paper, we leverage existing spike-based learning circuits to propose a biologically plausible architecture which, through forward and backward propagation of sensory input and error signals, can successfully classify distinct complex spatio-temporal spike patterns. Such a task, is non-trivial due to the nature of the spatio-temporal patterns. This new architecture relies solely on weight updates triggered by local variables and parameters, thus being suitable for implementation on mixed-signal analog/digital neuromorphic chips. This will enable the construction of low power always-on learning chips that can be applied to edge computing, robotics, and distributed computation. Due to the low-power nature of the hardware the network is implemented for, we expect it to be a good candidate for bio-signals processing~\cite{Donati_etal18b} and brain-machine interfaces~\cite{Corradi_Indiveri15}. Further work will include the detailed circuit-level implementation and simulation of the network layout, including transient and Monte-Carlo simulations to fully validate the network's robustness to parameter mismatch and variability.

\section*{Acknowledgment}
The authors would like to thank Joao Sacramento for fruitful discussions and comments, Ole Richter, Felix Bauer and Melika Payvand for original ideas and insight about the circuit and equations. \\ \\
This work was partially supported by the Swiss National Science Foundation Sinergia project \#CRSII5-18O316 and the ERC Grant "Neu-roAgents" (724295).

\bibliography{biblio/biblioncs.bib}
\bibliographystyle{IEEEtran}
 
\end{document}